# Clinical-Protocol-Aware Cyber Resilience for Connected Clinical Technologies Under AI-Enabled Adversarial Conditions

Pranav Patel
*Systems Engineering & Operations Research*
*MediTechSafe, Inc.*
Cincinnati, USA

Siva Kannathasan
*Systems Engineering & Operations Research*
*MediTechSafe, Inc.*
Cincinnati, USA

Prerak Patel
*Systems Engineering & Operations Research*
*MediTechSafe, Inc.*
line 4: Cincinnati, USA

Jigar Patel
*Systems Engineering & Operations Research (Advisor)*
*MediTechSafe, Inc.*
Cincinnati, USA

Vansh Patel
*System Engineering & Operations Research*
*MediTechSafe, Inc.*
Cincinnati, USA

Sheel Patel
*Information Technology Support & Engineering*
*TeamLogic IT*
Cincinnati, USA

***Abstract*— Healthcare increasingly depends on connected medical devices, clinical information systems, cloud services, and distributed-care technologies whose cybersecurity consequences can extend across clinical workflows and organizational boundaries. AI-enabled capabilities can accelerate reconnaissance, vulnerability discovery, and attack-path analysis, increasing the importance of architectures that constrain compromise propagation. This paper proposes a clinical-protocol-aware cyber-resilience framework based on bounded clinical micro-networks and controlled interconnections. The framework distinguishes technical cyber exposure from modeled clinical consequence and evaluates prevention, containment, and incident response within a parameterized Monte Carlo simulation. Experiments examine architectural connectivity, adversarial capability, security controls, clinical protocol dependency, clinical–technical risk alignment, communication directionality, lifecycle integration, and response timing. Under the specified simulation assumptions, a connected architecture increases mean cyber exposure by 56.3% relative to a segmented architecture, while combined prevention and containment reduce mean cyber exposure by 60.1% relative to a no-control scenario. Increasing protocol dependency raises the modeled clinical-impact index from 1.021 to 7.567 while mean cyber exposure remains unchanged, demonstrating that technical exposure and clinical consequence can diverge under the model's clinical weighting scheme. Risk-alignment experiments further show higher modeled clinical impact when greater technical risk corresponds to greater clinical significance. These results support comparative architectural analysis, not calibrated estimates of real-world cyber risk, control effectiveness, or patient harm; the adversarial-capability parameter does not quantify a particular AI system. The principal contribution is a clinical-context-aware approach that integrates technical exposure with clinical relevance, protocol dependency, patient-facing potential, and information dependencies to inform security-boundary design, control prioritization, and clinically coordinated containment across connected care environments.**

## I. Introduction

Healthcare increasingly depends on interconnected medical devices, electronic health record (EHR) systems, picture archiving and communication systems (PACS), laboratory information systems (LIS), clinical decision-support platforms, cloud services, and remote-monitoring technologies [1], [2]. These components exchange information across clinical workflows and organizational boundaries, creating cybersecurity dependencies that cannot be adequately assessed by examining individual devices in isolation. Moreover, patient-centered care spans home monitoring, primary care, diagnostic laboratories, imaging centers, hospitals, pharmacies, and rehabilitation services. Clinical protocols frequently depend on information generated in one setting being available, timely, accurate, and appropriately associated when consumed in another. Consequently, infrastructure ownership, operational responsibility, and cybersecurity practices may differ across a patient's care journey, although the resulting consequences can converge at the point of care.

A clinically relevant cybersecurity boundary therefore does not necessarily coincide with an organizational boundary, hospital, Internet Protocol (IP) subnet, virtual local area network (VLAN), or physical location. A medical device may be individually secure while the clinical ecosystem remains vulnerable because of relationships among devices, applications, users, privileges, supporting services, data flows, and clinical workflows [1], [16]. Conversely, compromise of a component without a direct patient-facing function may affect clinical operations if it can influence a clinical protocol. The 2021 Elekta cloud incident illustrates this dependency: disruption of an oncology information system interrupted radiation-treatment workflows across multiple healthcare organizations, demonstrating that clinical functions can become unavailable without direct compromise of the equipment performing them [5], [6]. This systemic perspective reinforces the need for lifecycle cybersecurity management and secure development of connected medical technologies [3], [4], [15].

Other incidents illustrate distinct risks to clinical safety and continuity. FDA actions concerning the Automated Impella Controller addressed vulnerabilities involving network or physical access that could potentially compromise device control or cause unexpected pump stoppage [7]; a separate software-related recall highlighted additional concerns about controller operation [8]. In 2026, FDA actions addressed a Portrait Core vulnerability that could enable privileged access and disrupt monitoring or alarms, a Centricity Universal Viewer flaw that could expose credentials and permit data manipulation, and an AW Server vulnerability affecting certain GE Revolution CT configurations [9]–[11]. Boston Scientific reported a cyber incident that disrupted new remote-monitoring activations for certain cardiac rhythm-management devices [18]. Separately, reporting on Epic Systems described AI-assisted identification of potential unauthorized-access and audit-trail weaknesses, raising concerns about possible patient-record alteration [19]. Although these incidents involve different failure modes, they illustrate cybersecurity dependencies affecting treatment continuity, device safety, monitoring, diagnostic availability, remote care, and clinical-information integrity.

Managing these dependencies is complicated by fragmented responsibilities across information technology, clinical engineering, clinical informatics, quality and compliance, healthcare providers, and technology vendors. Multi-vendor environments and evolving infrastructure further complicate lifecycle security management. Although established guidance addresses cybersecurity risk management and healthcare-sector security priorities [12]–[15], translating these principles into coordinated, clinically informed architectural decisions remains challenging. Conventional, siloed security engineering can prioritize technical vulnerabilities and network exposure while overlooking the clinical functions and protocols dependent on affected technologies. Consequently, architectures may reduce technical risk without adequately limiting clinical consequences. **Security engineering should therefore integrate clinical relevance, protocol dependency, and patient-facing functions into threat assessment, architectural design, lifecycle management, and incident response** to constrain compromise propagation and its potential clinical consequences.

AI-enabled adversarial capabilities further heighten these challenges by assisting reconnaissance, vulnerability analysis, entry-point discovery, and attack-path exploration. Resilient architectures must constrain both initial compromise and subsequent propagation while accounting for the clinical functions that cybersecurity management and incidents may affect.

This paper proposes a clinical-protocol-aware cybersecurity assessment and cyber-resilience framework for connected healthcare technologies under AI-enabled adversarial conditions. Adopting a cross-disciplinary systems-engineering perspective, the framework integrates clinical factors into threat assessment and modeling to inform secure-by-design architecture, lifecycle security management, and clinically informed incident response. Parameterized synthetic clinical environments and abstract adversarial-capability parameters enable controlled, reproducible experiments across diverse architectural and defensive configurations without requiring full healthcare-environment or AI-system emulation. The objective is to derive practical engineering principles for strengthening cybersecurity while accounting for clinical dependencies and potential clinical consequences.

The contributions are fivefold: 1) a clinical micro-network abstraction representing bounded clinical functions and their interconnections; 2) a clinically informed threat-assessment and modeling approach supporting secure-by-design and lifecycle security decisions; 3) distinct measures of compromise, cyber exposure, clinical-function disruption, and modeled clinical consequence; 4) architecture-aware evaluation of prevention, containment, communication directionality, and clinical dependencies; and 5) reproducible comparative experiments examining how architectural and defensive choices influence modeled resilience under varying adversarial conditions. The objective is to derive actionable engineering principles for limiting compromise propagation and reducing modeled clinical consequences across connected care environments.

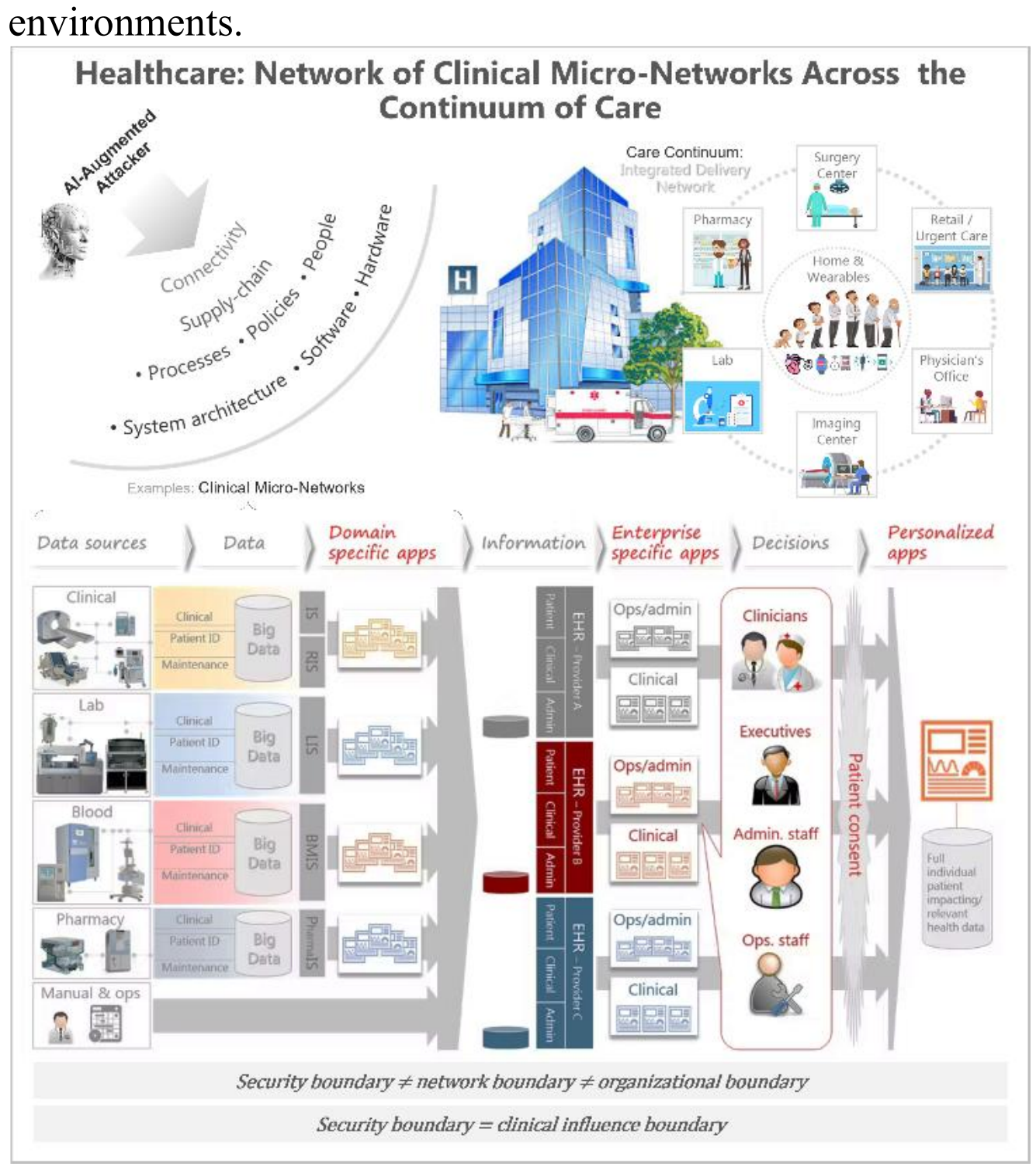

*Fig. 1. Clinical-protocol-aware cyber-resilience architecture*

*illustrating bounded clinical micro-networks, controlled interconnections, clinical information dependencies, and potential cross-network influence pathways across the continuum of care. The clinically relevant security boundary is defined by potential clinical influence rather than solely by network or organizational boundaries.*

## II. Clinical-Protocol-Aware Architecture

### A. Clinical Micro-Networks and Boundaries

Let the healthcare technology ecosystem be represented as

$$\boldsymbol{H} = \{\boldsymbol{M_1}, \boldsymbol{M_2}, \dots, \boldsymbol{M_n}\}$$

where each $M_i$ is a bounded collection of technologies supporting a clinical or operational function. Examples include patient monitoring, medication administration, imaging, laboratory testing, surgery, clinical applications, enterprise clinical IT, and distributed care.

A micro-network is a functional abstraction, not necessarily a physical subnet. It may span multiple devices, applications, and services connected through controlled interfaces. The architecture is represented as a directed influence graph $G = (V, E)$, where nodes represent technologies and edges represent potential pathways through which compromise could influence other components.

A patient-centered pathway may span home devices, gateways, cloud services, clinical applications, laboratories, pharmacies, and hospitals. Consequently, organizational ownership and network location do not fully define the clinical cyber-resilience boundary. The relevant boundary is the set of technologies and relationships through which a cyber event could influence a defined clinical function or protocol.

### B. Clinical Protocol Dependency and Information Continuity

A clinical protocol defines how clinicians, devices, applications, and information interact to achieve a clinical objective. The same technical vulnerability may therefore have different consequences depending on the affected workflow. A compromised administrative workstation and a compromised component supporting medication administration may have similar technical exposure but markedly different clinical significance.

We define a synthetic clinical-impact index to quantify modeled clinical-function disruption:

$$\boldsymbol{I}_{\mathbf{clin}} = \sum_{i \in V} \boldsymbol{D_i}\, \boldsymbol{r_i} \boldsymbol{d_i} \boldsymbol{f_i} \boldsymbol{k_i},$$

where $D_i \in \{0, 1\}$ indicates whether node $i$ experiences clinical-function disruption; $r_i$, $d_i$, and $f_i$ represent clinical relevance, protocol dependency, and patient-facing potential, respectively; and $k_i$ represents the node's criticality weight. The index is intended for comparative simulation. Clinical-function disruption is distinguished from technical compromise: a compromised node does not necessarily experience clinical-function disruption, and disruption is modeled probabilistically as described in Section IV.

Clinical information continuity is equally important. Information may become unavailable, delayed, altered, incomplete, incorrectly associated with a patient, or received without adequate provenance. Disruption of a laboratory system, integration engine, EHR, home-device gateway, or cloud service can therefore affect clinical decisions without directly compromising a patient-facing device. The consequences depend on the information involved, its role in the clinical protocol, and the availability of alternative sources or workflows.

Accordingly, the framework extends confidentiality, integrity, and availability analysis by explicitly relating information security to clinical dependencies. Auditability also matters: the reported Epic example [19] highlights how weaknesses in access logging and audit trails may impede investigation and accountability, even when an application remains operational.

## III. Architecture-Aware Resilience

### A. Prevention, Containment and Incident Response

Prevention, segmentation, containment, and incident response address different aspects of cyber resilience. Prevention reduces the probability of initial compromise. Segmentation constrains the pathways available for communication and cross-network propagation before an incident occurs. Containment limits further propagation after compromise, while incident response, for example, determines which affected components to isolate, when to act, and within what operational capacity. In healthcare, these measures must account for clinical dependencies and continuity-of-care requirements.

Protocol-aware response prioritizes nodes according to their modeled clinical relevance, protocol dependency, patient-facing potential, and criticality. However, prioritizing a node for isolation does not establish that immediate disconnection is clinically safe. Abrupt isolation may interrupt monitoring, treatment delivery, or information continuity, particularly during an active procedure. Safe response therefore requires coordination with clinical protocols, operational workarounds, and recovery procedures. The response model in this paper evaluates selection priority, capacity, and delay under specified assumptions; it does not represent a deployable clinical decision rule.

These mechanisms complement zero-trust principles [12], NIST CSF 2.0 [13], and healthcare-specific cybersecurity practices [14]. The proposed framework does not replace those approaches; it adds a clinical-function dimension to control selection and prioritization.

### B. Directionality and Lifecycle Change

Unidirectional information transfer and bidirectional influence are architecturally distinct. Reverse-direction pathways may allow an intermediary system to influence a clinical function through configuration, commands, or other interactions. Such pathways warrant security verification and validation proportionate to their potential clinical impact.

Healthcare technology architectures also change over time. A locally connected device may later acquire cloud connectivity, remote-management functions, or an interface to an EHR. These additions can change its clinical security context without changing the device itself. The framework therefore supports a lifecycle sequence of design, verification, validation, deployment, monitoring, change, and re-verification or re-validation, consistent with medical-device lifecycle guidance [3], [4], [15]. Security assessments should be revisited when interfaces, privileges, communication direction, or clinical dependencies change.

## IV. System Model And Mathematical Formulation

All model parameters, including those governing network structure, compromise and propagation probabilities, adversarial capability, defensive controls, and clinical-function disruption, are experimental assumptions rather than empirically calibrated estimates of real-world conditions. They are selected to represent bounded probabilities and specified qualitative relationships, enabling reproducible comparisons of architectural and defensive configurations.

### *A. Network Representation and Node Attributes*

The healthcare technology environment is represented by a directed graph $G = (V, E)$, where $V$ is the set of synthetic technology nodes and $E$ is the set of directed influence edges. The baseline configuration contains $N = |V| = 100$ nodes partitioned into five clinical micro-networks. Each node represents a technology or service supporting a clinical or operational function. An edge $(i, j) \in E$ represents a modeled pathway through which compromise of node $i$may propagate to node $j$.

Each node $i$ is assigned technical attributes $v_i, a_i, q_i, t_i, k_i$, representing vulnerability, accessibility, privilege, trust, and criticality, respectively. Clinical attributes $r_i, d_i, f_i$ represent clinical relevance, protocol dependency, and patient-facing potential. The attributes are synthetic and normalized to $[0, 1]$, unless otherwise specified.

Within-micro-network edge-generation probability is $0.06$. Cross-micro-network edge-generation probabilities are $0.006$ for the segmented architecture and $0.025$ for the connected architecture. These values define the baseline architectural comparison and are model assumptions.

For each generated edge, its influence weight is defined as

$$\boldsymbol{w_{ij} = 0.3q_i + 0.3t_i + 0.2a_j + 0.2u_{ij}},$$

where $u_{ij} \sim U(0, 1)$ is an independent random contribution used during edge generation.

### *B. Clinical-Technical Risk Alignment*

Clinical–technical risk alignment describes the correspondence between a technology's technical compromise potential and the clinical significance of the functions it supports. An aligned configuration associates greater technical risk with nodes of greater clinical significance, whereas a misaligned configuration places relatively greater technical risk on nodes of lower clinical significance. Alignment therefore concerns the correspondence between two risk dimensions, not whether a system is secure or insecure. Under comparable aggregate cyber exposure, compromise concentrated on clinically relevant, protocol-dependent, or patient-facing functions can produce greater modeled clinical consequences than compromise concentrated on less clinically consequential functions. In this study, alignment is evaluated by comparing paired configurations under the specified risk-assignment procedure; the resulting differences are conditional on the model's attributes and disruption mechanism.

### *C. Initial Compromise and Propagation*

Let $A, P, K, S \in [0, 1]$ denote AI-enabled adversarial capability, prevention strength, containment strength, and segmentation strength, respectively. The parameter $A$ represents a capability level reflecting accelerated analysis of complex environments, discovery of accessible entry points, and potential expansion through connected systems. Initial compromise probability for node $i$ is

$$\boldsymbol{p_i^{(0)} = \mathrm{clip}[(0.08 + 0.20v_i + 0.10a_i) \times (1 + 0.80A) \times (1 - 0.75P), 0, 0.95]},$$

where $\mathrm{clip}(x, l, u) = \min\big(u, \max(l, x)\big)$. Initial compromise is sampled using a Bernoulli distribution.

For an eligible directed edge $(i, j)$, the probability of propagation during round $r$ is

$$\boldsymbol{p_{ij}^{(r)} = \mathrm{clip}[\,(0.025 + 0.18w_{ij} + 0.18v_j) \times (1 + 1.25A)b_{ij}(S)(1 - 0.65K), 0, 0.95]},$$

where $b_{ij}(S) = 1$ for within-micro-network edges and $b_{ij}(S) = 1 - 0.85S$ for cross-micro-network edges. Propagation is simulated for three rounds. Successful attempts add previously uncompromised targets to the cumulative compromised set. Containment reduces propagation probability, while segmentation preferentially restricts cross-micro-network propagation.

### *D. Cyber Exposure and Clinical-Function Disruption*

Let $C_i \in \{0, 1\}$ indicate whether node $i$ is compromised at the end of the simulation. Cyber exposure is defined as

$$\boldsymbol{E_{\mathrm{cyber}} = \sum_{i \in V} C_i\, k_i(0.5 + 0.5q_i)}.$$

This measure combines compromise status, node criticality, and privilege, reflecting factors considered in attack-graph-based cybersecurity risk analysis. Conditional on compromise, clinical-function disruption is modeled as

$$\boldsymbol{D_i \mid C_i = 1 \sim \mathrm{Bernoulli}\big(p_i^{(D)}\big)},$$

where

$$\boldsymbol{p_i^{(D)} = \mathrm{clip}(0.05 + 0.20r_i + 0.20d_i + 0.10f_i, 0, 0.90)}.$$

Here, $r_i$, $d_i$, and $f_i$ denote clinical relevance, protocol dependency, and patient-facing potential, respectively. For uncompromised nodes, $D_i = 0$. Given the specified attribute bounds, the resulting disruption probability lies between 0.05 and 0.55. The clinical-impact index is

$$\boldsymbol{I_{\mathrm{clin}} = \sum_{i \in V} D_i\, r_i d_i f_i k_i}.$$

The model therefore distinguishes technical compromise, measured by $E_{\mathrm{cyber}}$, clinical-function disruption, represented by the disrupted-node count, and clinical consequence $I_{\mathrm{clin}}$.

*E. Monte Carlo Estimation*

For $T = 100$ Monte Carlo trials, the estimated mean of outcome $Y$ is

$$\bar{Y} = \frac{1}{T}\sum_{t=1}^{T} Y_t .$$

An approximate 95% confidence interval for the Monte Carlo mean is

$$\bar{Y} \pm 1.96\frac{s_Y}{\sqrt{T}},$$

where $s_Y$ is the sample standard deviation across trials. These intervals quantify Monte Carlo uncertainty conditional on the modeled setup. Any assessment of variation across independently generated network structures must be reported separately. The simulation uses a fixed random seed of 42 for reproducibility. The number of trials, graph-generation procedure, random-number handling, and parameter values must remain consistent with the published implementation.

*F. Incident Response and Protocol-Aware Isolation*

The response model compares three strategies: no additional response, generic response, and protocol-aware response. Under the generic strategy, eligible compromised nodes are ordered by ascending node identifier. Under the protocol-aware strategy, they are ordered by decreasing clinical-priority score,

$$\boldsymbol{Q_i = r_i d_i f_i k_i}.$$

Let $B$ denote response capacity, measured as the maximum number of nodes selected for isolation during a response round, and let $\tau$ denote the response-delay round. When a response is configured, the eligible compromised nodes are ranked and up to $B$ nodes are isolated before propagation in round $\tau$. Isolation prevents further propagation through the isolated node according to the modeled isolation rule. Previously isolated nodes are not selected again by the same response process.

The principal response experiment uses $B = 5$ and $\tau = 1$. Sensitivity analysis examines $B \in \{2, 5, 10\}$ and $\tau \in \{0, 1, 2\}$. Outcomes are compared using cyber exposure, disrupted-node count, and the clinical-impact index.

The response model isolates the effects of selection priority, response capacity, and delay under specified assumptions. It does not represent clinical triage, device recovery, operational workarounds, or the full consequences of isolating a technology during active care. Protocol-aware prioritization therefore does not establish that immediate isolation is clinically safe; actual response planning must account for monitoring, treatment delivery, information continuity, and clinically appropriate alternatives.

## V. Simulation

The simulation is a transparent comparative experiment. The implementation uses Python with NumPy, pandas, and Matplotlib. The principal settings are summarized below.

| Parameter | Value |
|---|---|
| Random seed | 42 |
| Monte Carlo trials | 100 |
| Synthetic nodes | 100 |
| Clinical micro-networks | 5 |
| Propagation rounds | 3 |
| Independent network replicates (robustness experiment) | 20 |
| Clinical-placement replicates per scenario | 20 |
| Risk-alignment replicates per scenario | 20 |
| AI capability levels | 0.0–1.0 (6 steps) |
| Prevention levels | 0.0–1.0 (6 steps) |
| Containment levels | 0.0–1.0 (6 steps) |
| Segmentation parameter (scenario based) | 0.0–1.0 |
| Protocol dependency | 0.0–1.0 |

The experiments evaluate architecture, AI-enabled adversarial capability, prevention and containment, clinical protocol dependency, clinical–technical risk alignment, communication directionality, lifecycle integration, multi-network robustness, and incident-response capacity and delay.

## VI. Experiments And Results

*A. Architecture and AI-Enabled Adversarial Capability*

TABLE I

Architectural Comparison

| Metric | Segmented | Connected |
|---|---|---|
| Cyber exposure, mean | 14.462 | 22.603 |
| Clinical-impact index, mean | 2.567 | 3.787 |
| Compromised nodes, mean | 35.52 | 55.59 |
| Disrupted nodes, mean | 13.22 | 20.16 |
| 95% CI: cyber exposure | [13.845, 15.080] | [21.896, 23.309] |
| 95% CI: clinical impact | [2.377,2.757] | [3.549,4.025] |

With the node inventory held constant, the connected configuration increased mean cyber exposure from 14.462 to 22.603 (56.3%) and the clinical-impact index from 2.567 to 3.787 (47.5%). Mean compromised nodes increased from 35.52 to 55.59, and mean disrupted nodes increased from 13.22 to 20.16. These results show that architectural connectivity materially affects modeled cyber exposure and clinical consequences.

Increasing modeled AI-enabled adversarial capability from 0.0 to 1.0 raised mean cyber exposure from 12.884 to 29.333, the clinical-impact index from 2.268 to 4.849, compromised nodes from 31.97 to 72.74, and disrupted nodes from 11.85 to 26.25. The experiment illustrates the architectural importance of limiting unnecessary connectivity and restricting the potential reach of compromise as AI accelerates reconnaissance, vulnerability discovery, entry-point identification, and attack-path analysis.

*B. Prevention and Containment*

TABLE II

Security-Control Effects

| Scenario | Cyber Exposure | Clinical Impact | Compromised Nodes | Disrupted Nodes |
|---|---|---|---|---|
| Neither | 32.309 | 5.681 | 79.72 | 29.98 |
| Prevention | 21.735 | 3.834 | 53.64 | 20.12 |
| Containment | 24.178 | 4.298 | 59.88 | 22.38 |
| Combined | 12.903 | 2.089 | 31.97 | 11.81 |

Relative to the no-control scenario, prevention alone reduced mean cyber exposure by 32.7%, containment alone by 25.2%, and combined controls by 60.1%. Combined controls also reduced the clinical-impact index by 63.2%, compromised nodes by 59.9%, and disrupted nodes by 60.6%. Under the evaluated settings, combining prevention and containment produced the largest reductions across all four outcomes.

The separate graded-control sweep showed decreasing modeled exposure and clinical impact as control strength increased. At control level 1.0, mean cyber exposure was 15.833 for prevention, 21.423 for containment, and 6.924 for combined controls, compared with 32.344 at level 0.0. These results belong to a separate experiment configuration and should not be conflated with the above table.

### C. *Clinical Protocol Dependency*

TABLE III

PROTOCOL-DEPENDENCY EFFECTS

| Protocol Dependency | Disrupted Nodes | Clinical Impact |
|---|---|---|
| 0.2 | 18.480 | 1.021 |
| 0.5 | 22.050 | 3.011 |
| 0.8 | 26.030 | 5.611 |
| 1.0 | 28.360 | 7.567 |

Mean cyber exposure remained at 25.526 across the four protocol-dependency settings, while the modeled clinical-impact index increased from 1.021 at dependency 0.2 to 7.567 at dependency 1.0. Mean disrupted-node counts increased from 18.48 to 28.36.

This divergence demonstrates that unchanged technical exposure does not imply unchanged clinical consequences. Conventional security assessments focused on technical vulnerabilities and network exposure may not adequately capture clinical protocol dependencies. Therefore, cybersecurity assessments should explicitly incorporate clinical dependencies alongside technical exposure during architectural design and lifecycle management.

### D. *Clinical-Function Placement and Risk Alignment*

TABLE IV

RISK-ALIGNMENT OUTCOMES

| Outcome | Aligned | Misaligned | Paired Difference (95% CI) |
|---|---|---|---|
| Clinical impact | 4.732 | 3.794 | 0.938 [0.895, 0.981] |
| Disrupted nodes | 24.228 | 22.423 | 1.805 [1.654, 1.956] |

Two experiments examined the relationship between clinical attributes and technical risk. The clinical-attribute permutation experiment, which held node criticality constant, produced no statistically significant mean change in the clinical-impact index (mean paired difference = −0.023; 95% CI [−0.125, 0.078]). This result does not establish that clinical-function placement is unimportant; rather, the experiment did not demonstrate a clear average effect under the specified permutation procedure and disruption model. In the separate risk-alignment experiment, aligned configurations produced a higher mean clinical-impact index than misaligned configurations (4.732 versus 3.794; paired difference = 0.938, 95% CI [0.895, 0.981]; $t(19) = 45.85$, $p < 0.001$). Mean disrupted-node counts were also higher (24.228 versus 22.423; paired difference = 1.805, 95% CI [1.654, 1.956]; $t(19) = 24.96$, $p < 0.001$), while mean cyber exposure remained identical at 25.807.

These experiments address different questions. Permuting clinical attributes tests whether their reassignment alone produces a detectable average change, whereas risk alignment explicitly tests the consequences of associating greater technical exposure with greater clinical significance. The latter provides clearer evidence, within this model, that technical-risk prioritization and clinical significance should be considered together when designing healthcare cybersecurity architectures. These findings are conditional on the modeled assumptions.

### E. *Communication Directionality and Lifecycle Integration*

In the directionality experiment, the bidirectional configuration increased cyber exposure from 25.709 to 35.947 and the clinical-impact index from 4.437 to 6.331 relative to the unidirectional configuration. Compromised nodes increased from 63.33 to 87.78, and disrupted nodes increased from 23.76 to 32.72. These results reflect the additional influence pathways represented in the bidirectional configuration.

TABLE V

LIFECYCLE-INTEGRATION EFFECTS

| Configuration | Edges | Cyber Exposure | Clinical Impact |
|---|---|---|---|
| Segmented baseline | 167 | 16.610 | 2.835 |
| Integrated, unidirectional | 202 | 17.332 | 3.029 |
| Integrated, bidirectional | 396 | 24.916 | 4.097 |

Adding 35 unidirectional integration edges increased mean cyber exposure from 16.610 to 17.332 and clinical impact from 2.835 to 3.029 relative to the segmented baseline. The bidirectional integrated configuration contained 396 edges and produced cyber exposure of 24.916 and clinical impact of 4.097, increases of approximately 50.0% and 44.5%, respectively, relative to baseline. These results support reassessing security boundaries when cloud connectivity, remote management, or clinical-system interfaces are introduced.

### F. *Multi-Network Robustness*

TABLE VI

ROBUSTNESS

| Architecture | Exposure (Mean ± SD) | Clinical Impact (Mean ± SD) |
|---|---|---|
| Connected | 21.369 ± 1.516 | 3.383 ± 0.333 |
| Segmented | 14.651 ± 1.055 | 2.380 ± 0.220 |

Across the multi-network robustness experiment, segmentation reduced mean cyber exposure from 21.369 to 14.651 and the clinical-impact index from 3.383 to 2.380. Mean compromised nodes decreased from 52.033 to 35.662, and disrupted nodes from 18.611 to 12.875. The segmented configuration also showed lower reported standard deviations across these outcomes, supporting the direction of the principal architectural finding across the evaluated network replicates.

### *G. Incident-Response Timing*

The response-capacity and delay sweep showed that intervention timing affected modeled outcomes. At response capacity 10 and delay 0, protocol-aware response yielded mean cyber exposure of 19.699 and a clinical-impact index of 3.412. At delay 2, these increased to 24.868 and 4.231, respectively, consistent with continued propagation during delayed containment. The model does not capture the clinical hazards of disconnecting devices during active care; the findings therefore support timely, clinically coordinated containment, not indiscriminate isolation or a deployable response policy.

## VII. Discussion And Engineering Implications

The results support a clinical-context-aware approach to healthcare cyber resilience under AI-enabled adversarial conditions. Within the model, segmentation reduced cyber exposure, combined prevention and containment produced the largest reductions among evaluated control scenarios, and clinical protocol dependency and risk alignment influenced clinical outcomes beyond aggregate cyber exposure. These findings motivate architectures and risk assessments that complement technical security parameters with explicit clinical factors, including clinical relevance, protocol dependency, patient-facing potential, and clinical-function criticality, to evaluate not only compromise propagation but also its potential consequences for clinical functions and information continuity.

Connected healthcare environments span devices, clinical applications, cloud services, and distributed-care platforms with differing ownership, privileges, and operational constraints. AI-enabled adversaries can accelerate reconnaissance, vulnerability discovery, and attack-path analysis across these heterogeneous environments, potentially exposing pathways between supporting services and clinically significant functions. These findings support constraining unnecessary connectivity and limiting compromise propagation rather than relying exclusively on preventing initial compromise.

The findings suggest five practical priorities:

1. **Define security boundaries and assess risk by clinical function and influence.** Map the devices, applications, gateways, and supporting services required for a clinical workflow, including dependencies that cross organizational and network boundaries. Assess conventional technical attributes, including vulnerability, accessibility, and privilege, alongside clinical relevance, protocol dependency, patient-facing potential, and the consequences of disrupting each function. For medication administration, for example, this may include medication systems, patient-identification services, the EHR, and interfaces supplying medication or allergy information. Determine which interfaces can change clinical data, issue commands, modify configuration, or interrupt required information flows. Use these dependencies to define security boundaries and restrict unnecessary cross-network access.
2. **Combine prevention with containment.** Combined controls produced the largest reductions among the evaluated scenarios. Reduce opportunities for initial compromise while limiting privileges, cross-network pathways, and propagation after compromise, particularly where supporting services can influence multiple clinical functions.
3. **Protect information continuity and integrity.** Clinical resilience depends on more than device availability. Delayed, altered, incomplete, or incorrectly associated information can affect clinical decisions even when patient-facing devices remain operational. Controls should preserve trustworthy information flows, provenance, and auditability across device, application, and cloud interfaces.
4. **Reassess security throughout the lifecycle.** New cloud connections, remote-management functions, EHR interfaces, or bidirectional pathways can change clinical security dependencies. Reassess affected boundaries and verify and validate relevant changes.
5. **Coordinate containment with clinical operations.** Isolation can introduce additional hazards if it interrupts monitoring, treatment, or essential information flows. Response plans should distinguish connections that can be blocked immediately from those requiring clinical coordination and identify safe degraded modes, alternative information sources, manual workarounds, and recovery procedures. Protocol-aware prioritization can inform response design, but the present model does not establish a clinically safe isolation policy.

For example, the security significance of a medication-administration interface depends on both its technical vulnerabilities and its potential influence on patient identification, medication orders, allergy information, and verification workflows. Assessment should account for these dependencies, restrict unnecessary access, and establish alternative verification and recovery procedures where isolation could interrupt care. The relevant security boundary follows the interface's potential influence on the clinical workflow, not merely its network location or organizational ownership.

The framework supports assessment within clinical micro-networks, across their interfaces, and along care pathways spanning home, clinic, hospital, laboratory, pharmacy, and remote-care services. At each level, technical exposure should be assessed alongside clinical relevance, protocol dependency, patient-facing potential, and information continuity to inform architectural design, containment, and recovery.

## VIII. Limitations And Future Work

The model represents a synthetic healthcare technology environment rather than a specific hospital, manufacturer, or clinical workflow. Its vulnerability, accessibility, trust, criticality, protocol-dependency, and adversarial-capability parameters are not empirically calibrated. It does not explicitly represent patient condition, clinical staff, redundancy, downtime procedures, organizational behavior, device recovery, or the full clinical consequences of isolation.

The results support comparative analysis under specified assumptions; they are not validated estimates of attack probability, real-world control effectiveness, or patient harm. Confidence intervals and paired significance tests quantify uncertainty within the evaluated experimental design, not external validity.

Future work should examine representative healthcare architectures and incorporate empirically supported attack pathways, temporal dependencies, redundancy, clinical workarounds, protocol-specific failure modes, and recovery behavior. Validation against appropriately documented operational data would be required before using the framework to estimate real-world risk or clinical consequences.

## IX. Conclusion

Healthcare cybersecurity spans connected devices, clinical applications, cloud services, and care pathways crossing technical and organizational boundaries. AI-enabled capabilities can accelerate reconnaissance, vulnerability discovery, and attack-path analysis, strengthening the case for architectures that constrain the reach of successful compromise. This paper proposed a clinical-protocol-aware framework using bounded clinical micro-networks and controlled interconnections to distinguish technical cyber exposure from modeled clinical consequence.

Under the specified simulation assumptions, segmentation and combined prevention and containment reduced modeled exposure, while protocol dependency, risk alignment, communication directionality, and response timing influenced comparative clinical outcomes. The central engineering contribution is to complement technical security assessment with explicit clinical factors, including clinical relevance, protocol dependency, patient-facing potential, and information dependencies, when designing cybersecurity architecture and response strategies. By combining technical controls with clinical coordination and continuity requirements, the framework provides a basis for evaluating cyber resilience across connected healthcare environments as adversarial capabilities evolve.